%% file: conference_101719.tex
\documentclass[conference]{IEEEtran}
\IEEEoverridecommandlockouts

\usepackage{amsmath}
\usepackage{bm}
\usepackage{listings}
\newcommand{\CenterRow}[2]{
  \dimen0=\ht\strutbox%
  \advance\dimen0\dp\strutbox%
  \multiply\dimen0 by#1%
  \divide\dimen0 by2%
  \advance\dimen0 by-.5\normalbaselineskip%
  \raisebox{-\dimen0}[0pt][0pt]{#2}}

\usepackage{cite}
\usepackage{amsmath,amssymb,amsfonts}
\usepackage{algorithmic}
\usepackage{graphicx}
\usepackage{textcomp}
\usepackage{xcolor}
\usepackage{tikz}
\def\BibTeX{{\rm B\kern-.05em{\sc i\kern-.025em b}\kern-.08em
    T\kern-.1667em\lower.7ex\hbox{E}\kern-.125emX}}

\newcommand\copyrighttext{%
  \footnotesize \textcopyright 2024 IEEE. Personal use of this material is permitted. Permission from IEEE must be obtained for all other uses, including reprinting/republishing this material for advertising or promotional purposes, collecting new collected works for resale or redistribution to servers or lists, or reuse of any copyrighted component of this work in other works.}

\newcommand\copyrightnotice{%
\begin{tikzpicture}[remember picture,overlay]
\node[anchor=south,yshift=10pt] at (current page.south) {\fbox{\parbox{\dimexpr0.75\textwidth-\fboxsep-\fboxrule\relax}{\copyrighttext}}};
\end{tikzpicture}%
}

\renewcommand\fbox{\fcolorbox{red}{white}}
\begin{document}

\title{Effects of Gabor Filters on \\Classification Performance of CNNs Trained on \\a Limited Number of Conditions}

\author{\IEEEauthorblockN{1\textsuperscript{st} Akito Morita}
\IEEEauthorblockA{\textit{Graduate School of Information Science and Technology} \\
\textit{Osaka Institute of Technology}\\
Osaka, Japan \\
d1d22a02@oit.ac.jp}
\and
\IEEEauthorblockN{2\textsuperscript{nd} Hirotsugu Okuno}
\IEEEauthorblockA{\textit{Faculty of Information Science and Technology} \\
\textit{Osaka Institute of Technology,}\\
Osaka, Japan \\
hirotsugu.okuno@oit.ac.jp}
}
\maketitle
\copyrightnotice
\begin{abstract}
In this study, we propose a technique to improve the accuracy and reduce the size of convolutional neural networks (CNNs) running on edge devices for real-world robot vision applications. CNNs running on edge devices must have a small architecture, and CNNs for robot vision applications involving on-site object recognition must be able to be trained efficiently to identify specific visual targets from data obtained under a limited variation of conditions. The visual nervous system (VNS) is a good example that meets the above requirements because it learns from few visual experiences. Therefore, we used a Gabor filter, a model of the feature extractor of the VNS, as a preprocessor for CNNs to investigate the accuracy of the CNNs trained with small amounts of data. To evaluate how well CNNs trained on image data acquired under a limited variation of conditions generalize to data acquired under other conditions, we created an image dataset consisting of images acquired from different camera positions, and investigated the accuracy of the CNNs that trained using images acquired at a certain distance. The results were compared after training on multiple CNN architectures with and without Gabor filters as preprocessing. The results showed that preprocessing with Gabor filters improves the generalization performance of CNNs and contributes to reducing the size of CNNs.
\end{abstract}


\input{Section1}
\input{Section2}
\input{Section3}
\input{Section4}

\input{Section5}
\input{Section6}

\bibliographystyle{IEEEtran}
\bibliography{references,references2}

\end{document}

%% file: Section1.tex
\section{Introduction}
The use of deep convolutional neural networks (CNNs) has become a promising option for a variety of image-based tasks over the past decade. Its applications have extended to edge devices, such as vision systems for autonomous mobile robots operating in real-world environments. However, edge devices have limited computational resources, and therefore, CNNs with small-scale architectures are desired rather than deep CNNs.

In addition, robot vision applications involving on-site object recognition require the identification of specific visual targets required for the application, and CNNs used in such situations must be able to efficiently learn such targets from a small amount of data. In other words, CNNs must be able to generalize to the appearance variations that occur in the real-world environment. In a real-world environment, the appearance of an object changes depending on a variety of factors, including the angle and distance between the object and the vision sensor, as well as lighting conditions. These factors change the shape, color and contrast of the projected object.

One solution to the generalizability problem is to enrich the training data. However, preparing training data that includes sufficient changes in the environment requires a large amount of effort, and therefore, it is desirable to learn the necessary features from a small amount of data.

On the other hand, the visual nervous system (VNS) of humans and animals is capable of learning from a small amount of visual experience. In addition, the VNS uses limited computational resources to perform visual perception with far less energy than today's computer systems. The VNS processes images in a hierarchical structure, extracting basic information such as orientation, color, and direction of motion in the early stages. Orientation selectivity modeled by a set of Gabor filters, which have been identified in a wide range of mammalian primary visual cortex, is expressed from an early stage of development with little visual experience \cite{espinosa_development_2012}, and therefore, acquisition of higher-order functions such as object recognition is likely to be supported by such preprocessing. For these reasons, the use of Gabor filters as a preprocessor for CNNs can be a promising approach to improve generalization performance in small-scale CNNs. Several studies have attempted to apply Gabor filters as a preprocessor for CNNs, suggesting that their use can improve classification accuracy \cite{akito_2022,taghi_2019}.

In the field known as neuromorphic engineering, attempts have been made to efficiently implement processing in the VNS, mainly utilizing analog VLSIs (aVLSIs), resulting in the development of many aVLSIs that simulate the characteristics of the retina and the primary visual cortex (\cite{delbruck_silicon_2004,shimonomura_multichip_2005} for examples) since the development of the pioneering silicon retina \cite{mead_silicon_1988,mead_neuromorphic_1990}. Since some neuromorphic devices provide spatial Gabor-like characteristics with low power consumption, combining such devices and a small-scale CNN is an effective way to achieve edge artificial intelligence (AI) with generalization performance in a small size and low power consumption.

The purpose of the study is to investigate to what extent the Gabor filter as a preprocessor contributes to the accuracy of CNNs trained on a limited number of conditions. This will allow us to evaluate the generalization performance of CNNs to different appearances of the target object. For this purpose, an image dataset was created in which images were acquired from different camera positions. This dataset was used to train and evaluate ResNet18 \cite{ResNet}, MobileNetV2 \cite{MobileNetV2}, and AlexNet \cite{AlexNet}, which have relatively small architectures among widely known CNN architectures. We also investigated the potential for Gabor filters to contribute to further miniaturization of these architectures.

%% file: Section2.tex
\section{Image Classification Algorithm}
\subsection{Processing Flow}
\begin{figure}[tb]
    \begin{center}
        \includegraphics[width=\columnwidth]{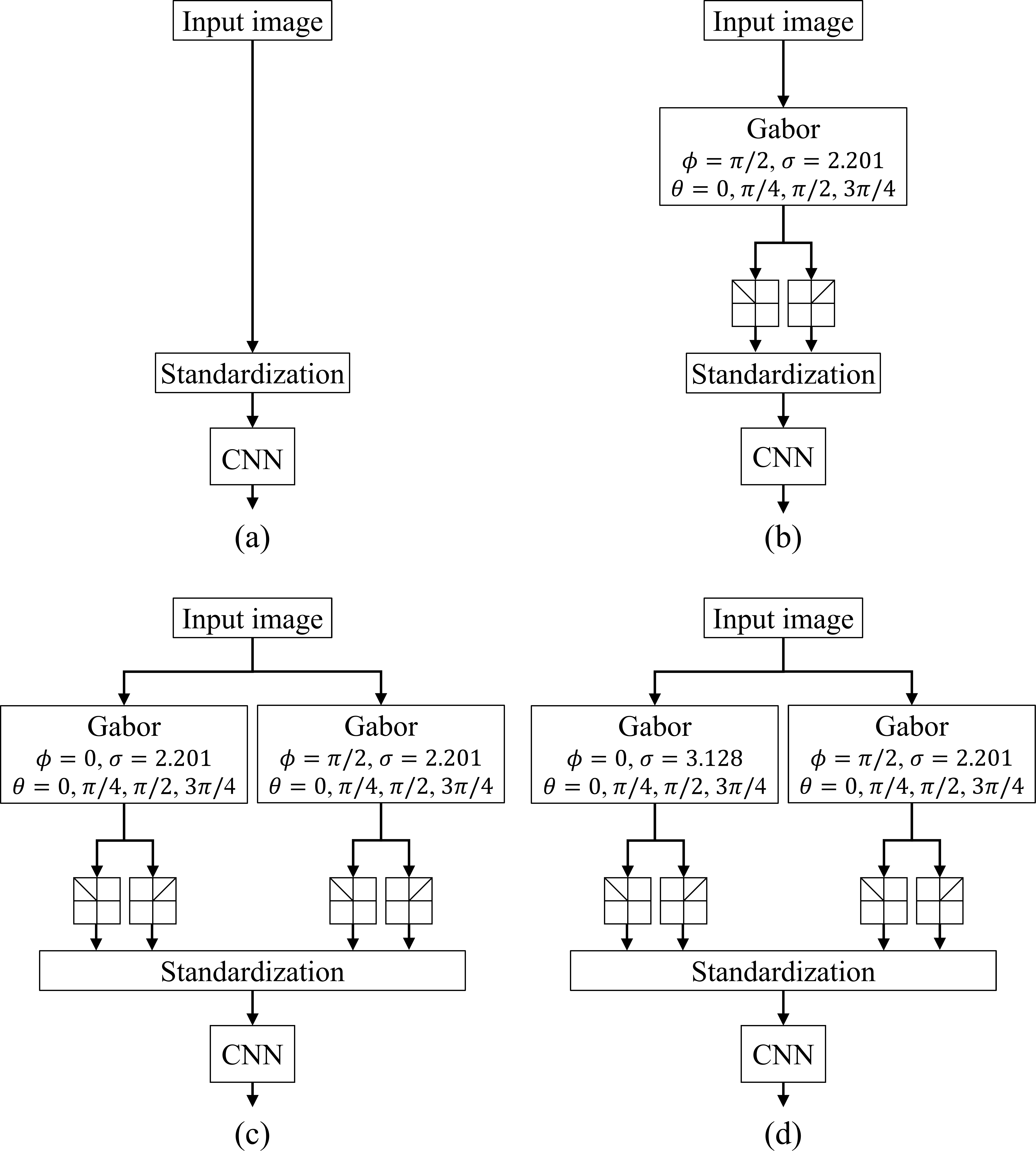}
    \caption{\label{Fig1} Processing flow. (a) Processing flow without Gabor filters. (b)-(d) Processing flow with Gabor filters. The rectangles below Gabor filters represent rectification.}
    \end{center}    
\end{figure}

We compared the classification results of CNNs without Gabor filters (Fig. \ref{Fig1}(a)) and CNNs that use preprocessing with Gabor filters (Fig. \ref{Fig1}(b)-(d)) in order to investigate the effects of Gabor filters as preprocessing for CNNs. The processes depicted in Fig. \ref{Fig1}(b)-(d) each use a combination of Gabor filters with different parameters. The process in Fig. \ref{Fig1}(c) uses two sets of Gabor filters whose phases are orthogonal to each other. The process in Fig. 1(d) uses two sets of Gabor filters whose phases are orthogonal to each other and whose scales are also different. The parameters are described in the next section. The input data is an 8-bit grayscale image. CNNs without Gabor filters receive images that are standardized only. On the other hand, CNNs that use Gabor filters receives images that have been subjected to the following three processes: Gabor filtering, rectification to positive and negative signals, and standardization.
\begin{figure}[tb]
    \begin{center}
        \includegraphics[width=.7\columnwidth]{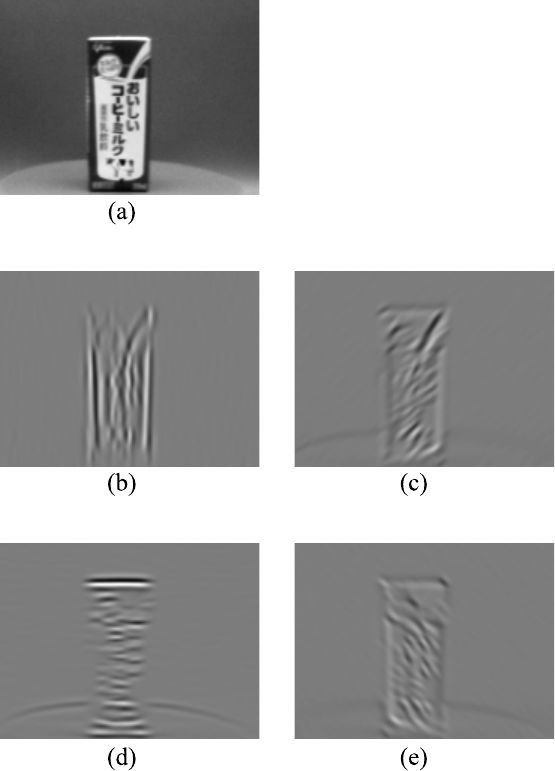}
    \caption{\label{Fig2} Examples of Gabor filtered images. (a) Input image. (b)-(e) Gabor filtered images. (b), (c), (d), and (e) are results for $\theta = 0, \pi/4, \pi/2, 3\pi/4$, respectively.}
    \end{center}    
\end{figure}
\begin{table}[b]
    \caption{\label{tbl: gabor parameters} Parameters of The Gabor Filters}
	\centering
	\begin{tabular}{c|c|c|c}
		\hline
		&\multicolumn{3}{c}{Parameters}\\
		\hline
        Methods&$\sigma$&$\lambda$&$\phi$\\
        \hline
		Fig. 1(b)&2.201&5.66&0\\
		\hline
		\CenterRow{2}{Fig. 1(c)}&2.201&5.66&0\\
                                &2.201&5.66&$\pi/2$\\
		\hline
		\CenterRow{2}{Fig. 1(d)}&3.128&8&0\\
                                &2.201&5.66&$\pi/2$\\
		\hline
	\end{tabular}
\end{table}

\subsection{Gabor Filters}
A Gabor filter is a two-dimensional spatial filter that enhances features with a particular frequency, orientation, and phase. The filter was used to simulate the spatial characteristics of a simple cell, which is a well-studied neuron in the primary visual cortex \cite{simple_cell}. The kernel of a Gabor filter that enhances edges with $\theta$ degree orientation is expressed as:
\begin{align}
    \label{eq: gabor}
    G(x, y) = &A \exp(- \frac{x^{2} + y^{2}}{2 \sigma^{2}}) \nonumber\\
                &\times \cos(\frac{2\pi}{\lambda} (x \cos\theta + y \sin\theta) - \phi ),
\end{align}
where $(x, y)$ represents the coordinate in the kernel. In this study, we used Gabor filters whose $\theta=0, \pi/4, \pi/2, 3\pi/4$. The parameters used in this study are listed in Table \ref{tbl: gabor parameters}.

Fig. \ref{Fig2} shows examples of Gabor filtered images. Fig. \ref{Fig2}(a) shows the input image, and (b), (c), (d), and (e) are Gabor filtered images with $\theta = 0, \pi/4, \pi/2, 3\pi/4$, respectively. In each image, features of a particular orientation are enhanced.

%% file: Section3.tex
\section{Dataset}
\label{section: datasets}
\begin{figure}[p]
    \centering
    \includegraphics[width=\columnwidth]{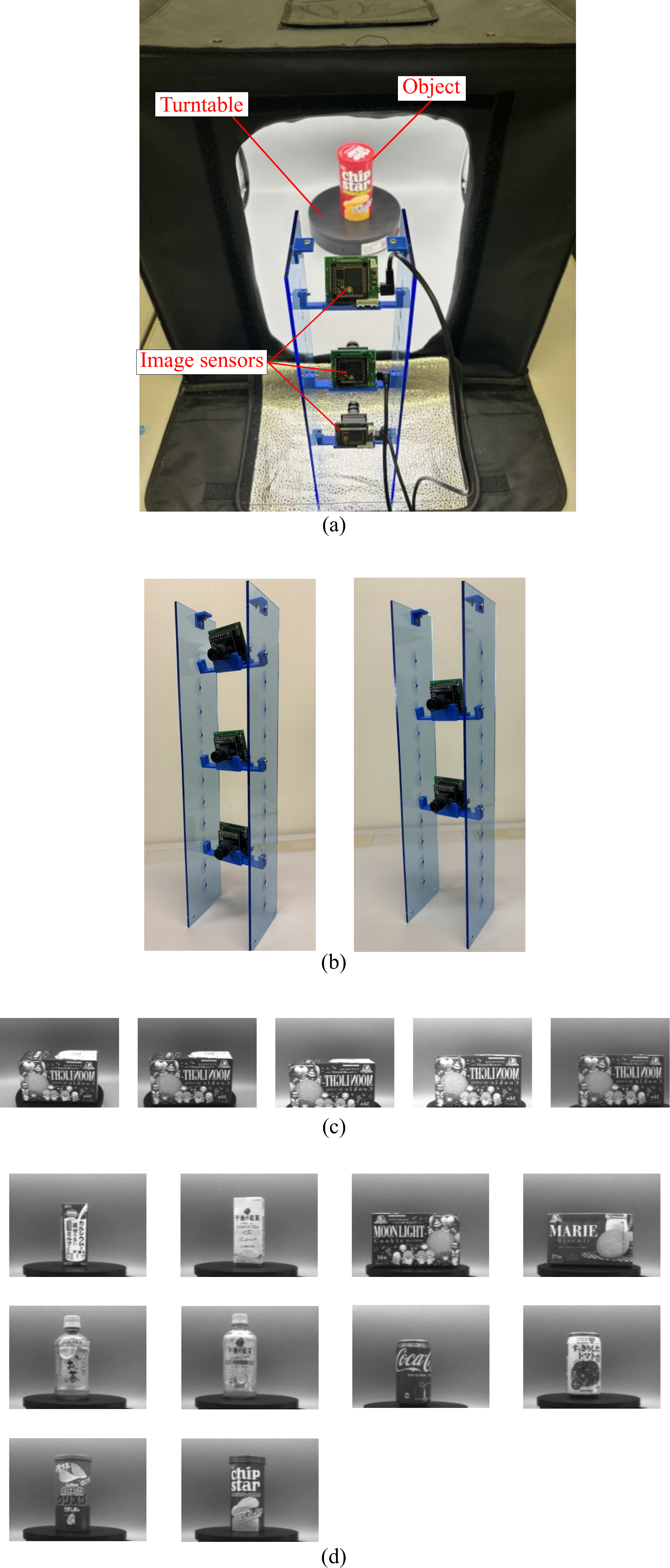}
    \caption{\label{Fig3} Environment for image acquisition and example images of the dataset. (a) Environment for image acquisition. (b) The frame used for camera height adjustment. (c) Examples of images taken from five different heights using the frame shown in (b). The heights of the camera positions are 34.0, 28.0, 22.0, 16.0, and 10.0 cm from the left image. (d) Example images of 10 objects in the dataset.}
\end{figure}

We created an image dataset to investigate the effect of Gabor filters on the classification accuracy of a CNN trained on images acquired under a limited variation of conditions. Fig. \ref{Fig3}(a) shows the environment in which the images for the dataset were acquired. The image acquisition was performed in a simple darkroom. The resolution of the image is $160 \times 120$ pixels. The dataset contained 10 different objects, and images of each object were acquired from four different distances (39.5, 47.0, 54.5, 62.0 cm from the turntable) and five different heights. 

For the height adjustment, we created the frame shown in Fig. \ref{Fig3}(b) from acrylic board by laser cutting. The distance between the height adjustment holes is 3.0 cm, and the height was set in 6.0 cm increments. Examples of images taken at each of the five heights are shown in Fig. \ref{Fig3}(c). 

For each condition, a turntable was used to acquire images from a 360-degree orientation. The total number of images in this dataset is 8400. Examples of the images in the dataset are shown in Fig. \ref{Fig3}(d).

%% file: Section4.tex
\section{Experimental Setups}
\subsection{Evaluation Method}
To evaluate the impact of preprocessing using Gabor filters on the classification accuracy of CNNs trained on images acquired under a limited number conditions, we investigated the accuracy using the dataset described in section \ref{section: datasets}. Only images acquired at certain distances were used as training data, while images acquired at all other distances were used as test data.

\subsection{CNN Architecture and Parameters}
We used ResNet18 \cite{ResNet}, MobileNetV2 \cite{MobileNetV2}, and AlexNet \cite{AlexNet} as CNN architecture.
The batch size used during training is 64 and the number of epochs is 50. Stochastic gradient descent with Nesterov momentum \cite{SGD_with_Nesterov} with a learning rate of $10^{-2}$, weight decay of $10^{-2}$, and momentum of 0.9 was used as the optimization method. The learning rate was warmed up for the first 5 epochs, after which the learning rate decayed according to the cosine decay scheduler \cite{cosine_decay_scheduler} without restarting.

We used Python v.3.11, PyTorch v.2.1 and Torchvision v.0.16 for the CNN implementation.

\subsection{Data Augmentations}
For data augmentation, the following operations were performed on the input image during training: random horizontal flipping, random cropping of the image to $110 \times 110$ pixels, and resizing to $224 \times 224$ pixels. During testing, the center of the image, $120 \times 120$ pixels, was resized to $224 \times 224$ pixels and used as input.

%% file: Section5.tex
\section{Results}
\subsection{Comparison of Accuracy}
\label{subsection: comparison of accuracy}
Table \ref{tbl: accuracy ResNet}, Table \ref{tbl: accuracy MobileNet}, and Table \ref{tbl: accuracy AlexNet} show the classification accuracy obtained by training with images acquired at each of four different distances. Each classification accuracy is the average of five trials trained with initial weights varied by a random number. Table \ref{tbl: accuracy ResNet} shows the accuracy of ResNet18, Table \ref{tbl: accuracy MobileNet} shows the accuracy of MobileNetV2, and Table \ref{tbl: accuracy AlexNet} shows the accuracy of AlexNet.

As can be seen from Table \ref{tbl: accuracy ResNet} and Table \ref{tbl: accuracy MobileNet}, no significant differences were observed between the cases with and without preprocessing when ResNet18 and MobileNetV2 were used. However, the method using multi-scale Gabor filters shown in Fig. \ref{Fig1}(c) showed higher accuracy on average.

On the other hand, as can be seen from Table \ref{tbl: accuracy AlexNet}, all the methods using the Gabor filters shown in Fig. \ref{Fig1}(b)-(d) as a preprocessing step obtained higher accuracy on average than those without preprocessing when the AlexNet, which has a small number of layers, is used. In particular, the method in which multi-scale Gabor filters are used showed the highest accuracy.

\subsection{Evaluation of Feature Vectors Represented by Each Layer}
The comparison of classification results of AlexNet shown in section \ref{subsection: comparison of accuracy} suggests that preprocessing with Gabor filters can be more effective on architectures with fewer layers. To confirm this, we investigated the discriminability of feature vectors generated by each layer of ResNet18 trained on images acquired at a distance of 54.5 cm, where there was no significant difference between the results with and without preprocessing. 

\begin{figure}[!h]
    \begin{center}
        \includegraphics[width=\columnwidth]{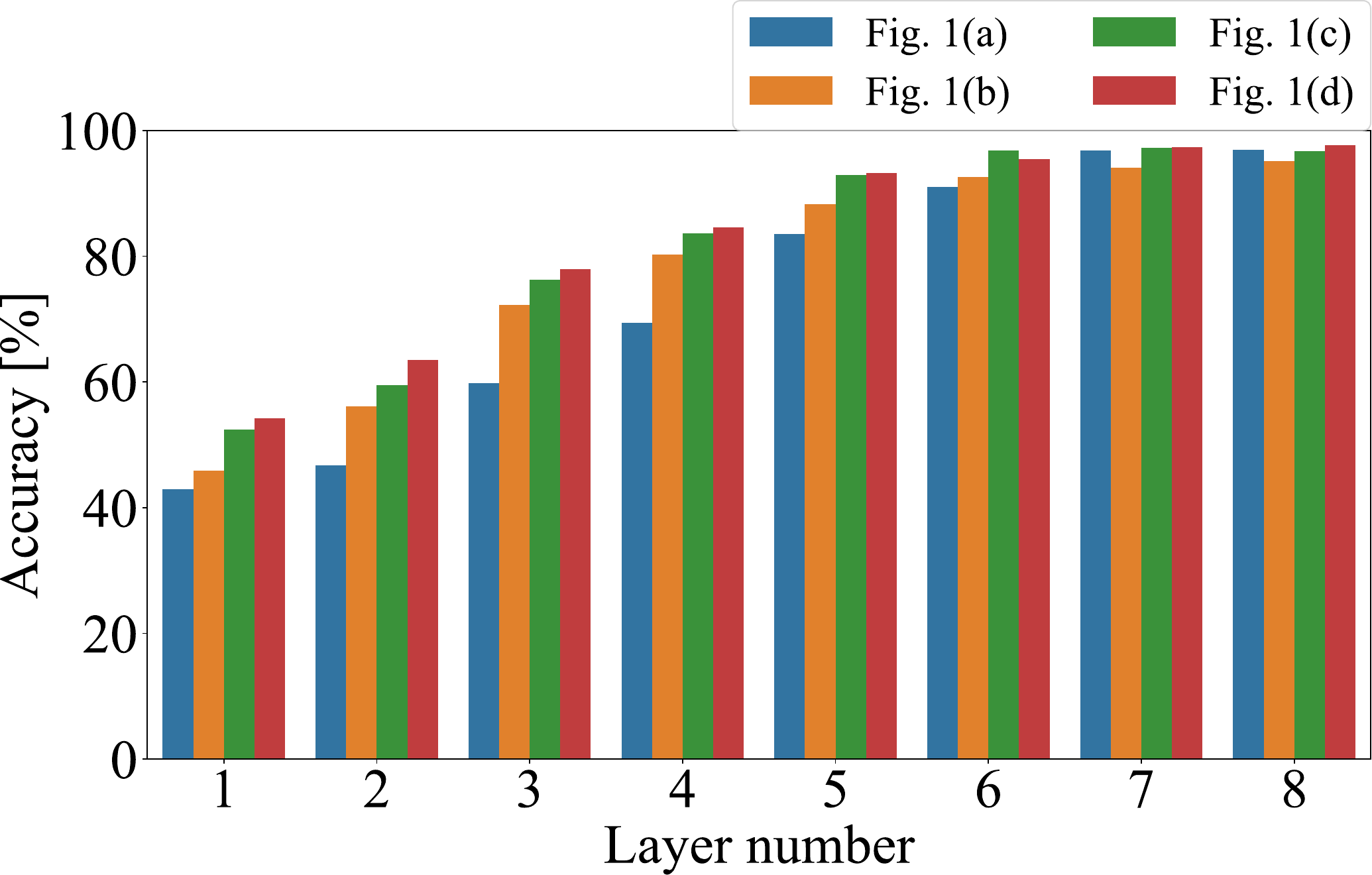}
    \caption{\label{fig: svm_result} Classification accuracy of linear SVMs that use feature vectors generated by each residual block of the ResNet18 as input. The layer number indicates the depth of the layer, where 1 indicates the first residual block and 8 indicates the last residual block.}
    \end{center}    
\end{figure}

For this purpose, we investigated the classification accuracy of a linear support vector machine (SVM) that uses feature vectors generated by each residual block of the trained ResNet18 as input. A Python function of a machine learning library (LiearSVC of scikit-learn v.1.4) was used to implement SVMs. The parameters used are the followings:
\begin{lstlisting}[basicstyle=\ttfamily\footnotesize]
LinearSVC(penalty='l2', loss='squared_hinge',
        dual='auto', tol=0.0001, C=1.0, 
        multi_class='ovr', intercept_scaling=1, 
        max_iter=1000)
 \end{lstlisting}

Fig. \ref{fig: svm_result} shows the linear SVMs classification accuracy at each residual block of the ResNet18 trained on images acquired at a distance of 54.5 cm. This figure shows that the classification accuracy peaks at an earlier layer when Gabor filters are used as preprocessing. This change is particularly pronounced for methods using multi-scale Gabor filters. The results suggest that the use of Gabor filters offers higher classification accuracy with smaller CNN architectures.

\begin{table}[tb]
    \centering
    \caption{Test Accuracy (\%) for ResNet18}
    \label{tbl: accuracy ResNet}
    \begin{tabular}{c|cccc|c}
        \hline
        \bf{Method}&\multicolumn{4}{|c|}{\bf{Camera positions used for training dataset}}\\
        \hline
        &39.5 cm& 47.0 cm& 54.5 cm& 62.0 cm& Average\\
        \hline
         Fig. 1(a)& 90.93& 83.47& 96.85& 96.97& 92.06\\
         Fig. 1(b)& 80.10& 77.83& 97.38& 97.64& 88.24\\
         Fig. 1(c)& 85.34& 83.49& 97.39& 98.12& 91.08\\
         Fig. 1(d)& 89.54& 91.36& 97.75& 98.90& 94.39\\
        \hline
    \end{tabular}
\end{table}
\begin{table}[tb]
    \centering
    \caption{Test Accuracy (\%) for MobileNetV2}
    \label{tbl: accuracy MobileNet}
    \begin{tabular}{c|cccc|c}
        \hline
        \bf{Method}&\multicolumn{4}{|c|}{\bf{Camera positions used for training dataset}}\\
        \hline
        &39.5 cm& 47.0 cm& 54.5 cm& 62.0 cm& Average\\
        \hline
         Fig. 1(a)& 74.68& 81.37& 96.51& 96.94& 87.37\\
         Fig. 1(b)& 72.47& 75.68& 98.85& 98.70& 86.43\\
         Fig. 1(c)& 73.71& 81.43& 96.64& 98.50& 87.57\\
         Fig. 1(d)& 79.13& 81.67& 96.14& 98.45& 88.85\\
        \hline
    \end{tabular}
\end{table}
\begin{table}[!h]
    \centering
    \caption{Test Accuracy (\%) for AlexNet}
    \label{tbl: accuracy AlexNet}
    \begin{tabular}{c|cccc|c}
        \hline
        \bf{Method}&\multicolumn{4}{|c|}{\bf{Camera positions used for training dataset}}\\
        \hline
        &39.5 cm& 47.0 cm& 54.5 cm& 62.0 cm& Average\\
        \hline
         Fig. 1(a)& 44.65& 48.96& 72.99& 81.09& 61.92\\
         Fig. 1(b)& 44.85& 59.15& 73.91& 86.26& 66.04\\
         Fig. 1(c)& 47.27& 56.47& 72.49& 89.07& 66.33\\
         Fig. 1(d)& 48.88& 57.12& 75.77& 92.62& 68.59\\
        \hline
    \end{tabular}
\end{table}

%% file: Section6.tex
\section{Conclusion}
In this study, we investigated to what extent the Gabor filter as a preprocessing for CNNs contributes to improving the accuracy and reducing the size of CNNs trained on data obtained under a limited number of conditions. 

The comparison of the classification accuracy showed that preprocessing with multi-scale Gabor filters is effective in improving, albeit slightly, the accuracy of CNNs trained on images acquired at a certain distance. This strength was also evident when using the AlexNet, which has a small number of layers. This result suggests that the use of Gabor filters could contribute to the improvement of generalization performance of CNNs. 

Experiments evaluating the classification accuracy of linear SVMs that use feature vectors generated by each residual block of the trained ResNet18 showed that feature vectors of the ResNet18 that use Gabor filters provide higher accuracy at shallower layers than feature vectors from CNNs that do not use Gabor filters. The results suggest that the use of Gabor filters contributes to the miniaturization of CNN architectures, and that the use of multi-phase and multi-scale Gabor filters further contributes to miniaturization.

Neurons that respond to features of various scales and various phases have been identified in the VNS \cite{ringachSpatialStructureSymmetry2002}, and simulating this property could help CNNs achieve the strengths of the VNS, i.e., learning from little visual experience and performing visual recognition in an energy efficient manner.

%% file: references.bib
@inproceedings{akito_2022,
  title = {The Effect of Preprocessing with {Gabor} Filters on Image Classification Using {CNNs}},
  booktitle = {The 2022 {{International Conference}} on {{Artificial Life}} and {{Robotics}}},
  author = {Morita, Akito and Okuno, Hirotsugu},
  year = {2022},
  month = jan,
  pages = {503--506},
  langid = {english}
}

@inproceedings{AlexNet,
  title = {{ImageNet} Classification with Deep Convolutional Neural Networks},
  booktitle = {Advances in {{Neural Information Processing Systems}}},
  author = {Krizhevsky, Alex and Sutskever, Ilya and Hinton, Geoffrey E},
  year = {2012},
  volume = {25},
  publisher = {Curran Associates, Inc.},
  urldate = {2024-04-18}
}

@inproceedings{cosine_decay_scheduler,
  title = {{{SGDR}}: {{Stochastic}} Gradient Descent with Warm Restarts},
  shorttitle = {{{SGDR}}},
  booktitle = {5th {{International Conference}} on {{Learning Representations}}},
  author = {Loshchilov, Ilya and Hutter, Frank},
  year = {2017},
  urldate = {2024-04-04}
}

@article{delbruck_silicon_2004,
  title = {A Silicon Early Visual System as a Model Animal},
  author = {Delbr{\"u}ck, Tobi and Liu, Shih-Chii},
  year = {2004},
  month = aug,
  journal = {Vision Research},
  volume = {44},
  number = {17},
  pages = {2083--2089},
  issn = {0042-6989},
  doi = {10.1016/j.visres.2004.03.021},
  urldate = {2024-03-28}
}

@article{espinosa_development_2012,
  title = {Development and Plasticity of the Primary Visual Cortex},
  author = {Espinosa, J. Sebastian and Stryker, Michael P.},
  year = {2012},
  month = jul,
  journal = {Neuron},
  volume = {75},
  number = {2},
  pages = {230--249},
  issn = {1097-4199},
  doi = {10.1016/j.neuron.2012.06.009},
  langid = {english},
  pmcid = {PMC3612584},
  pmid = {22841309}
}

@article{mead_neuromorphic_1990,
  title = {Neuromorphic Electronic Systems},
  author = {Mead, C.},
  year = {1990},
  month = oct,
  journal = {Proceedings of the IEEE},
  volume = {78},
  number = {10},
  pages = {1629--1636},
  issn = {1558-2256},
  doi = {10.1109/5.58356},
  urldate = {2024-03-28}
}

@article{mead_silicon_1988,
  title = {A Silicon Model of Early Visual Processing},
  author = {Mead, Carver A. and Mahowald, M. A.},
  year = {1988},
  month = jan,
  journal = {Neural Networks},
  volume = {1},
  number = {1},
  pages = {91--97},
  issn = {0893-6080},
  doi = {10.1016/0893-6080(88)90024-X},
  urldate = {2024-03-28}
}

@inproceedings{MobileNetV2,
  title = {{{MobileNetV2}}: {{Inverted Residuals}} and {{Linear Bottlenecks}}},
  shorttitle = {{{MobileNetV2}}},
  booktitle = {2018 {{IEEE}}/{{CVF Conference}} on {{Computer Vision}} and {{Pattern Recognition}}},
  author = {Sandler, Mark and Howard, Andrew and Zhu, Menglong and Zhmoginov, Andrey and Chen, Liang-Chieh},
  year = {2018},
  month = jun,
  pages = {4510--4520},
  doi = {10.1109/CVPR.2018.00474},
  urldate = {2024-04-19},
  isbn = {978-1-5386-6420-9},
  langid = {english}
}

@inproceedings{ResNet,
  title = {Deep Residual Learning for Image Recognition},
  booktitle = {2016 {{IEEE Conference}} on {{Computer Vision}} and {{Pattern Recognition}}},
  author = {He, Kaiming and Zhang, Xiangyu and Ren, Shaoqing and Sun, Jian},
  year = {2016},
  month=Jun,
  pages = {770--778},
  doi = {10.1109/CVPR.2016.90},
  urldate = {2024-03-11},
  isbn = {978-1-4673-8851-1},
  langid = {english}
}

@article{ringachSpatialStructureSymmetry2002,
  title = {Spatial Structure and Symmetry of Simple-Cell Receptive Fields in Macaque Primary Visual Cortex},
  author = {Ringach, Dario L.},
  year = {2002},
  month = jul,
  journal = {Journal of Neurophysiology},
  volume = {88},
  number = {1},
  pages = {455--463},
  issn = {0022-3077},
  doi = {10.1152/jn.2002.88.1.455},
  langid = {english},
  pmid = {12091567}
}

@inproceedings{SGD_with_Nesterov,
  title = {On the Importance of Initialization and Momentum in Deep Learning},
  booktitle = {Proceedings of the 30th {{International Conference}} on {{Machine Learning}}},
  author = {Sutskever, Ilya and Martens, James and Dahl, George and Hinton, Geoffrey},
  year = {2013},
  month = Jun,
  pages = {1139--1147},
  issn = {1938-7228},
  urldate = {2024-04-04},
  langid = {english}
}

@article{shimonomura_multichip_2005,
  title = {A Multichip {{aVLSI}} System Emulating Orientation Selectivity of Primary Visual Cortical Cells},
  author = {Shimonomura, K. and Yagi, T.},
  year = {2005},
  month = jul,
  journal = {IEEE Transactions on Neural Networks},
  volume = {16},
  number = {4},
  pages = {972--979},
  issn = {1941-0093},
  doi = {10.1109/TNN.2005.849845},
  urldate = {2024-03-28}
}

@article{simple_cell,
  title = {Mathematical Description of the Responses of Simple Cortical Cells},
  author = {Mar{\^c}elja, S.},
  year = {1980},
  month = nov,
  journal = {Journal of the Optical Society of America},
  volume = {70},
  number = {11},
  pages = {1297--1300},
  publisher = {Optica Publishing Group},
  doi = {10.1364/JOSA.70.001297},
  urldate = {2024-04-19},
  copyright = {{\copyright} 1980 Optical Society of America},
  langid = {english}
}

@article{taghi_2019,
  title = {Fast Facial Emotion Recognition Using Convolutional Neural Networks and {Gabor} Filters},
  author = {Taghi Zadeh, Milad Mohammad and Imani, Maryam and Majidi, Babak},
  year = {2019},
  month = feb,
  journal = {2019 5th Conference on Knowledge Based Engineering and Innovation},
  pages = {577--581},
  publisher = {IEEE},
  address = {Tehran, Iran},
  doi = {10.1109/KBEI.2019.8734943},
  urldate = {2024-03-28},
  copyright = {https://ieeexplore.ieee.org/Xplorehelp/downloads/license-information/IEEE.html},
  isbn = {9781728108728}
}
